\pdfoutput=1

\documentclass[11pt]{article}

\usepackage[preprint]{acl}

\usepackage{times}
\usepackage{latexsym}

\usepackage[T1]{fontenc}

\usepackage[utf8]{inputenc}

\usepackage{microtype}

\usepackage{inconsolata}

\usepackage{graphicx}

\usepackage{multirow}
\usepackage{amsmath}
\usepackage{booktabs}
\usepackage{colortbl}
\newcommand{\grayline}{\rowcolor[gray]{.90}}
\usepackage{enumitem}
\usepackage{xspace}
\usepackage{tcolorbox}
\usepackage{listings}
\usepackage{multicol}
\usepackage{lipsum}
\newcommand{\model}{CodeRM-8B\xspace}
\usepackage{subcaption}

\title{Dynamic Scaling of Unit Tests for Code Reward Modeling}

\author{
Zeyao Ma$^{1,3}$\thanks{Equal Contributions.}\thanks{Work was done when interned at Zhipu AI.},
Xiaokang Zhang$^{1,3}$\footnotemark[1],
Jing Zhang$^{1,3}$\thanks{Corresponding Author.},
Jifan Yu$^2$,
Sijia Luo$^1$,
Jie Tang$^2$
\\[3pt]
$^1$School of Information, Renmin University of China,
$^2$Tsinghua University,
\\
$^3$Key Laboratory of Data Engineering and Knowledge Engineering, Beijing, China
\\[1pt]
\small \texttt{zeyaoma@gmail.com, \{zhang2718, zhang-jing\}@ruc.edu.cn} \\[2pt]
\vspace{-4ex}
\url{https://code-reward-model.github.io}
}

\begin{document}
\maketitle

\begin{abstract}

Current large language models (LLMs) often struggle to produce accurate solutions on the first attempt for code generation.
Prior research tackles this challenge by generating multiple candidate solutions and validating them with LLM-generated unit tests.
The execution results of unit tests serve as reward signals to identify correct solutions.
As LLMs always confidently make mistakes, these unit tests are not reliable, thereby diminishing the quality of reward signals.
Motivated by the observation that scaling the number of solutions improves LLM performance, we explore the impact of scaling unit tests to enhance reward signal quality.
Our pioneer experiment reveals a positive correlation between the number of unit tests and reward signal quality, with greater benefits observed in more challenging problems.
Based on these insights, we propose \model, a lightweight yet effective unit test generator that enables efficient and high-quality unit test scaling.
Additionally, we implement a dynamic scaling mechanism that adapts the number of unit tests based on problem difficulty, further improving efficiency.
Experimental results show that our approach significantly improves performance across various models on three benchmarks (e.g., with gains of $18.43\%$ for Llama3-8B and $3.42\%$ for GPT-4o-mini on HumanEval Plus).

\end{abstract}

\section{Introduction}
\label{sec:introduction}
Code generation aims to automatically produce code solutions that satisfy programming requirements specified in natural language~\citep{codegen2020}.
Recent advancements in large language models (LLMs) have shown significant progress in this domain~\citep{touvron2023llama, gpt-4-report}.
However, generating correct code on the first attempt remains challenging due to the inherent complexity of reasoning required~\citep{li2022alphacode, huang2024large}.
Beyond developing more powerful LLMs, some research leverages additional test-time computation to generate more code solutions via repeated sampling, while a verifier or reward model reranks and identifies the optimal solution~\citep{inala2022faultaware, chen2023codet}.
Although repeated sampling enables LLMs to produce correct solutions~\citep{chen2021humaneval, brown2024monkey}, identifying the correct solution from the multitude of candidates poses a significant challenge~\citep{gao2023scalingrm}.


Unit tests (i.e., pairs of input and expected output) are frequently used as verifers to identify correct code solutions~\citep{shi2022mbr-exec, chen2023codet}.
Specifically, LLMs generate unit tests based on the given instructions and candidate code solutions, which are then executed by a compiler or interpreter.
The execution results serve as reward signals to identify correct solutions.
However, as LLMs often confidently make mistakes~\citep{huang2024hallu-survey}, the reliability of these unit tests is not guaranteed, thereby diminishing the quality of the reward signal.
Inspired by the performance gains from scaling test-time computation to generate more responses~\citep{snell2024scaling}, we ask:
\textit{Can generating more unit tests improve the quality of the reward signal for code solutions?}

To address this question, we conduct a pioneer experiment to investigate the correlation between the number of unit tests and the quality of the code reward signal across different LLMs, code solution quantities, and unit test scales.
Our findings reveal:

\begin{itemize}[noitemsep, topsep=0pt]
    \item Scaling the number of unit tests consistently improves the accuracy of identifying correct solutions across different model sizes and code solution quantities.
    \item The benefits of scaling unit tests depend on problem difficulty, with more computational resources yielding greater improvements for challenging problems.
\end{itemize}

Building on these observations, we develop \textbf{\model}, a small yet powerful unit test generator designed to facilitate efficient and high-quality unit test scaling.
To support model training, we introduce an automatic data synthetic pipeline that produces high-quality unit tests from existing code instruction-tuning datasets.
Leveraging this synthesized data, we perform supervised fine-tuning (SFT) on Llama3.1-8B, resulting in a high-quality unit test generation model.
Furthermore, since the benefits of scaling unit tests vary with problem difficulty, we follow \citet{damani2024learning} to implement the dynamic unit test scaling on different problems.
Specifically, we train a lightweight problem difficulty classifier using the language model probing method~\citep{Alain2017porb, kadavath2022language}, which extracts implicit information from the LLM's intermediate representation and outputs a scalar problem difficulty.
Based on this classifier, we dynamically allocate computation budgets across problems of varying difficulties using a greedy algorithm~\citep{edmonds1971matroids}.


We conduct extensive experiments to evaluate the effectiveness of \model on three widely used benchmarks and four LLMs with varying parameter scales for solution generation.
The results demonstrate that scaling unit tests with \model significantly improves the performance of smaller models (e.g., a performance gain of $18.43\%$ on HumanEval Plus for Llama3-8B).
Moreover, \model enhances the performance of significantly larger models or even proprietary models (e.g., a $4.95\%$ gain for Llama3-70B and $3.42\%$ for GPT-4o-mini on HumanEval Plus).
We also evaluate the performance of dynamic unit test scaling on two benchmarks.
By leveraging a trained problem difficulty classifier and dynamically allocating computation budgets, this approach brings additional performance improvements at a fixed computational cost (e.g., up to approximately $0.5\%$ performance gain on MBPP Plus).

The main contributions of this paper are as follows:
1) A pioneer experiment revealing a positive correlation between the number of unit tests and the quality of the code reward signal, with greater benefits for more challenging problems as unit test scales.
2) A small yet powerful model enabling efficient and high-quality unit test scaling.
3) An implementation of dynamic unit test scaling over problems in different difficulties.
4) Experimental results validating the effectiveness of \model and the implementation of dynamic scaling.

\section{Pioneer Experiment}
\label{sec:scaling}
This section explores the correlation between the quantity of LLM-generated unit tests and the quality of the code reward signal.
We first present the methodology, including a unit test-based majority voting framework and the setup of the pioneer experiment.
Subsequently, we analyze the observations derived from the pioneer results.

\begin{figure*}[t]
  \centering
  \includegraphics[width=\textwidth]{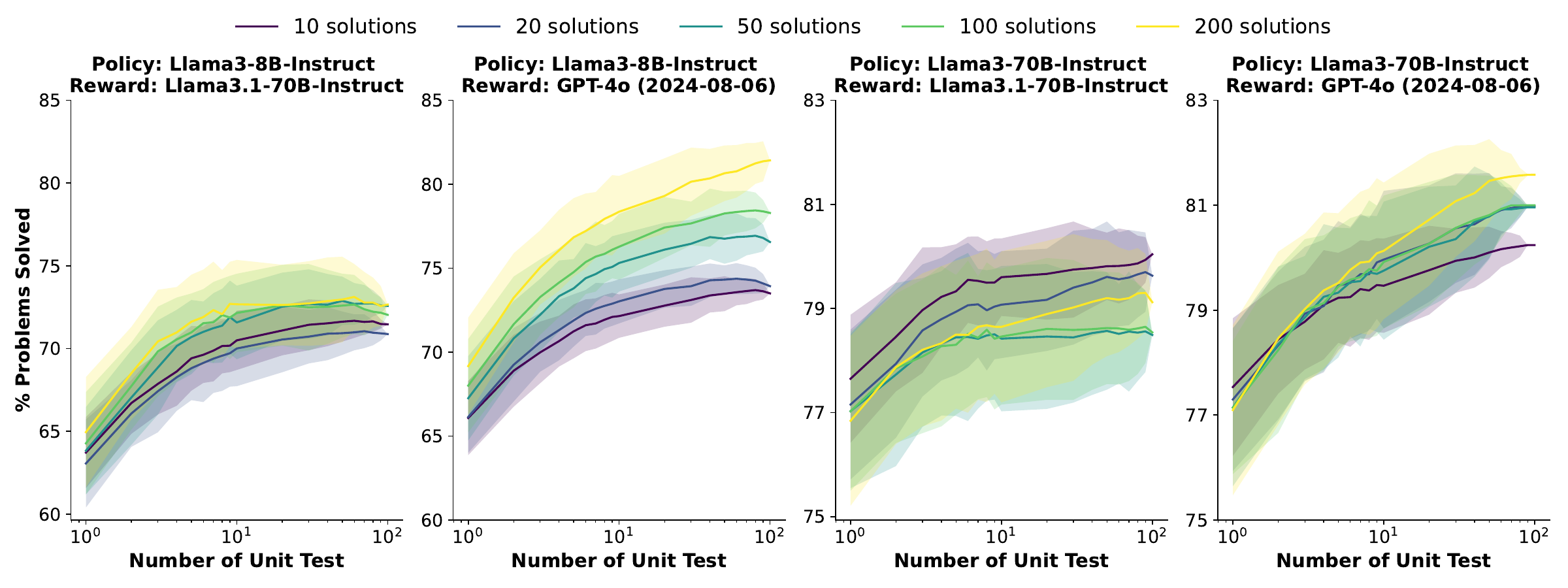}
  \caption{Scaling the quantities of unit tests for majority voting leads to improvements in performance across different policy models and reward models. Policy refers to the model that produces code solutions, while reward denotes the model that generates unit tests.}
 \label{fig:scale_on_diff_sol}
\end{figure*}

\begin{figure}[t]
  \centering
  \includegraphics[width=0.49\textwidth]{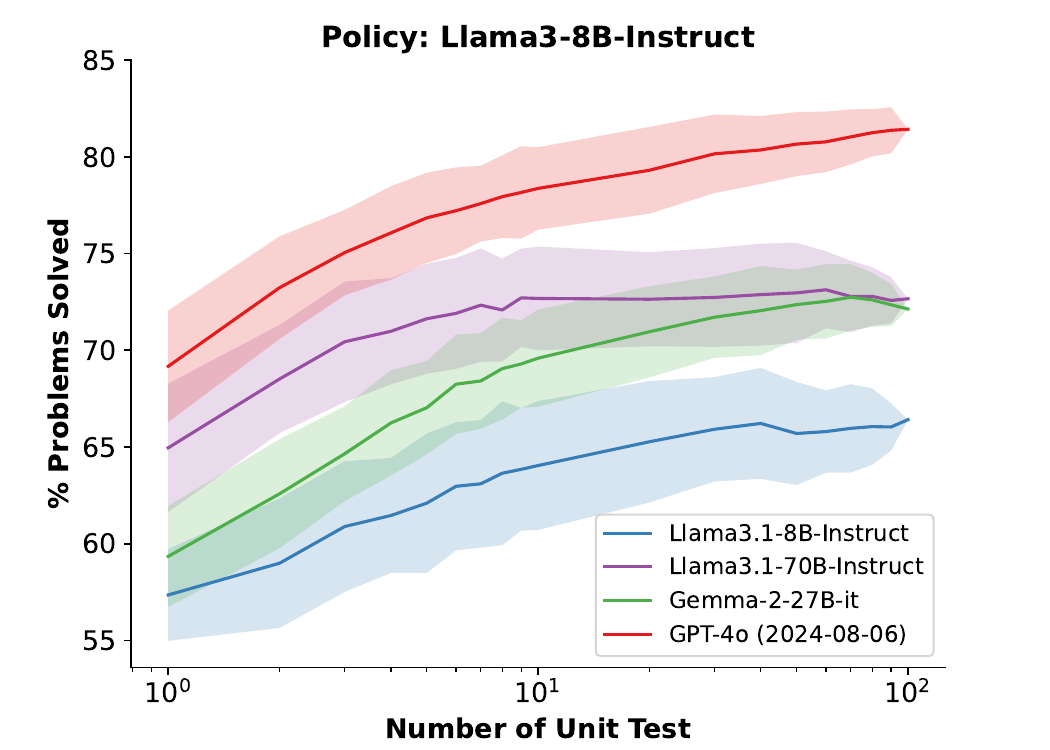}
  \caption{
  The correlation between the quantities of unit tests and the performance on different unit test generators (reward model) with $200$ candidate code solutions.
  }
 \label{fig:scale_on_diff_model}
\end{figure}

\subsection{Methodology}
\label{sec:methodology}

The unit test-based majority voting framework follows a standard best-of-N strategy~\citep{cobbe2021training, lightman2023let}.
Given a programming question $Q$, the policy model generates $N$ candidate code solutions:
\[
\{C_1, C_2, \dots, C_N\},
\]
where $C_i$ represents the $i$-th candidate solution.
Based on the programming question and the candidate code solution, an auxiliary LLM generates $M$ unit tests:
\[
\{T_1, T_2, \dots, T_M\}
\]
Each unit test $T_j$ consists of a set of test cases:
\[
T_j = \{(x_{j,1}, y_{j,1}), (x_{j,2}, y_{j,2}),\dots,\!(x_{j,K_j}, y_{j,K_j})\},
\]
where \( x_{j,k} \) is the input for the \( k \)-th test case in \( T_j \), and \( y_{j,k} \) is the corresponding expected output.
\( K_j \) represents the number of test cases in \( T_j \).
Each candidate solution \( C_i \) is executed on the unit tests \( \{T_1, T_2, \dots, T_M\} \). For a given unit test \( T_j \), the execution of \( C_i \) produces a binary result:
\[
r_{i,j} = 
\begin{cases} 
1, & \text{if \( C_i \) passes all test cases in \( T_j \)}; \\
0, & \text{otherwise}.
\end{cases}
\]
The binary results for all unit tests form a reward signal for each candidate solution:
\[
\mathbf{R}_i = \{r_{i,1}, r_{i,2}, \dots, r_{i,M}\}.
\]
Finally, we select the optimal candidate solution \( C_{\text{opt}} \) based on majority voting~\citep{wang2023selfconsistency}.
This voting process determines \( C_{\text{opt}} \) as the solution that passes the maximum number of unit tests:
\begin{equation}
C_{\text{opt}} = \underset{C_i}{\mathrm{argmax}} \sum_{j=1}^M r_{i,j}.
\label{eq:opt}
\end{equation}


We utilize HumanEval Plus~\citep{liu2023evalplus}, a widely adopted dataset comprising handwritten Python programming questions with comprehensive unit tests.
For each question, an LLM (policy model) generates $N = 200$ code solutions, while another LLM (reward model) produces $M = 100$ unit tests for supervision. 
The optimal solution is selected using Equation~\eqref{eq:opt} and validated against the ground truth unit tests in the dataset.
To compute the results for $n$ solutions ($0 \leq n \leq N$) and $m$ unit tests ($0 \leq m \leq M$), we employ the bootstrap resampling method~\citep{1976bootstrap}, generating $100$ bootstrap samples to compute mean values and confidence intervals, as shown in Figure~\ref{fig:scale_on_diff_sol} and~\ref{fig:scale_on_diff_model}.

\begin{figure*}[htbp]
  \centering
  \includegraphics[width=\textwidth]{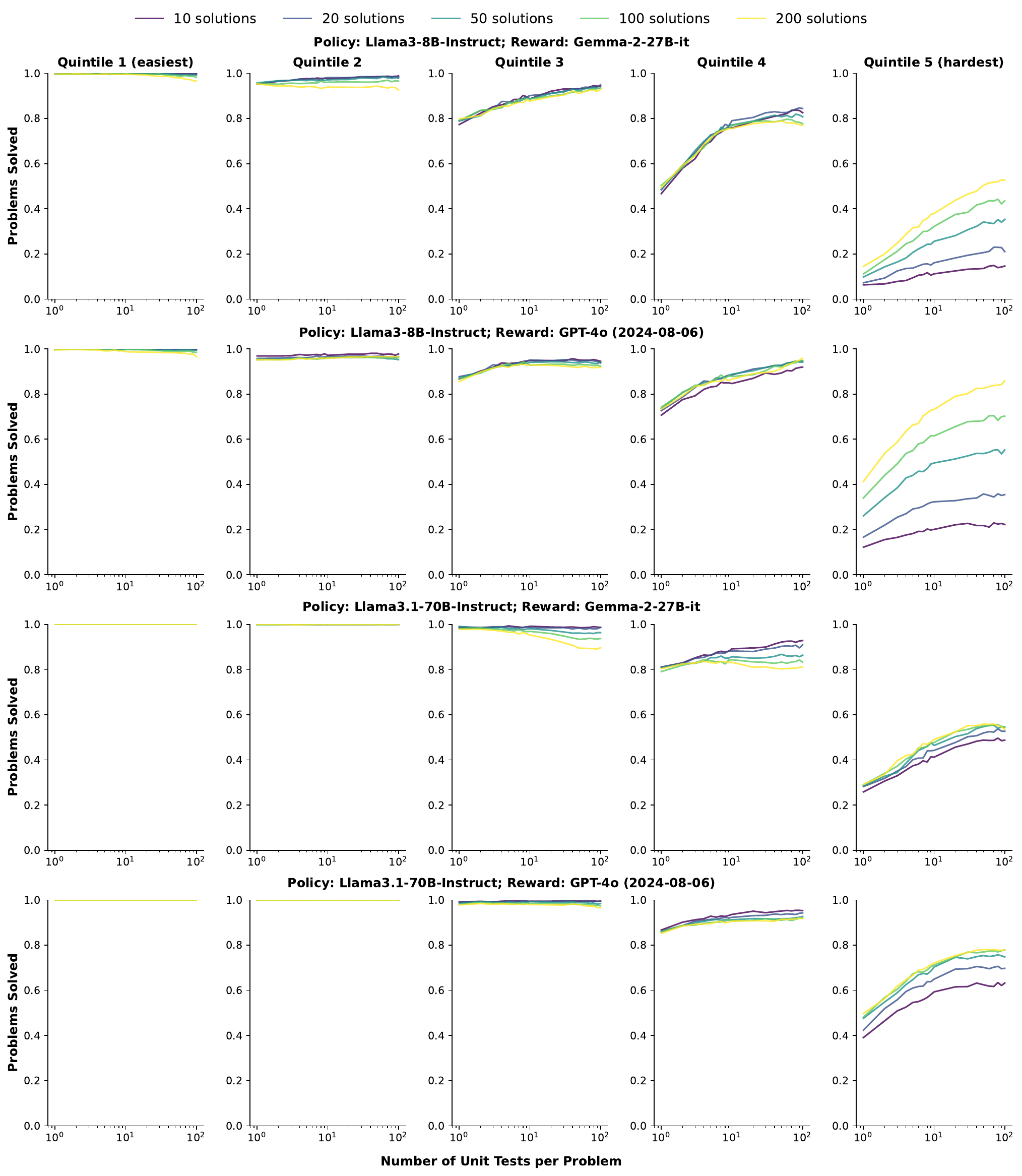}
  \caption{
  The improvements of best-of-N performance on problems of different difficulties.
  Quintile 1 (easiest) has the highest pass rate, while Quintile 2 (hardest) has the lowest pass rate.
  Scaling the quantity of unit tests significantly improves the accuracy on more complex problems.
  }
 \label{fig:scale_on_diff_diff}
\end{figure*}

\subsection{Observations}
\label{sec:observe}


\paragraph{Scaling the number of unit tests consistently improves the quality of the reward signal.}
Figure~\ref{fig:scale_on_diff_sol} demonstrates that increasing the number of unit tests consistently improves best-of-N performance across different quantities of code solutions, policy models, and reward models.
As the number of code solutions increases, performance typically improves with more samples, since a larger sample size enhances the likelihood of generating accurate responses~\citep{brown2024monkey}.
However, in the third sub-figure of Figure~\ref{fig:scale_on_diff_sol}, performance decreases with additional test-time computation.
This aligns with observations by~\citet{cobbe2021training}, where excessive test-time computation can generate adversarial solutions that mislead the verifier.
For different policy models, scaling unit tests yields more pronounced performance improvements for weaker models compared to stronger ones.
For instance, performance gains of $11\%$ and $5\%$ are observed for Llama3-8B and Llama3-70B, respectively, when employing GPT-4o as the reward model.

Figure~\ref{fig:scale_on_diff_model} compares the reward signals produced by different reward models.
Generally, models with more parameters achieve better best-of-N performance.
Notably, while Gemma-2-27B-it performs significantly worse than Llama3.1-70B with a single unit test, it achieves comparable performance when scaled to $100$ unit tests per question.
This may be attributed to smaller models generating responses with greater coverage and diversity, as discussed by \citet{bansal2024smaller}.


\paragraph{Scaling unit tests is more effective for harder problems.}
We evaluate the effectiveness of scaling unit tests across problems of different difficulty levels.
Specifically, we first eliminate problems without a single correct solution.
The remaining problems are divided into five equal parts based on the actual pass rate obtained via repeated sampling.
Figure~\ref{fig:scale_on_diff_diff} presents the results using Llama3-8B and Llama3.1-70B as the policy models, with Gemma-2-27B and GPT-4o as reward models.
The results demonstrate that the benefits of scaling unit tests are highly dependent on problem complexity.
For more challenging problems, increasing computational resources yields greater performance enhancement.
This highlights the potential of dynamically scaling of unit tests based on problem difficulty, representing a viable approach for resource conservation within a fixed computational budget.
More fine-grained results are provided in Appendix~\ref{app:more_results}.

\section{Towards Efficient and High-Quality Unit Test Scaling}
\label{sec:train}
\begin{figure*}[t]
  \centering
  \includegraphics[width=\textwidth]{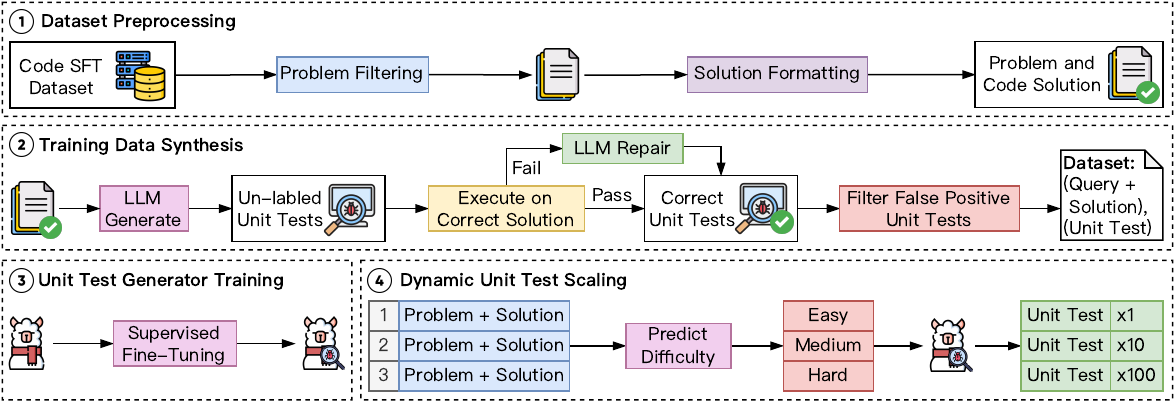}
  \caption{Overview for efficient and high-quality unit test scaling. First, we train a lightweight unit test generator based on high-quality synthetic data. Subsequently, we employ dynamic unit test scaling to further improve efficiency.}
 \label{fig:main}
\end{figure*}

In light of the observations in Section~\ref{sec:scaling}, we propose \model, a small yet effective unit test generator designed to enable efficient and high-quality unit test scaling.
To this end, we introduce a synthetic data pipeline for generating high-quality unit tests and model training.
Additionally, as shown in Section~\ref{sec:scaling}, scaling unit tests proves to be more effective for harder problems.
To further improve efficiency, we implement a dynamic scaling strategy that adapts to problems of varying difficulty.
Specifically, following \citet{damani2024learning}, we train a problem difficulty classifier and employ a greedy algorithm to allocate computational resources dynamically, prioritizing harder problems.

\subsection{Unit Test Generator}
To train an effective unit test generator, we first construct high-quality training data through a data synthetic pipeline that includes dataset preprocessing and unit test generation, as illustrated in the first two sections in Figure~\ref{fig:main}.
We then apply supervised fine-tuning (SFT) to optimize the generator.

\paragraph{Dataset Preprocessing.}
We utilize existing high-quality code instruction-tuning datasets as the foundation of our pipeline, including CodeFeedback-Filtered-Instruction\footnote{\url{https://huggingface.co/datasets/m-a-p/CodeFeedback-Filtered-Instruction}}~\citep{zheng2024opencodeinterpreter} and the training set of TACO\footnote{\url{https://huggingface.co/datasets/BAAI/TACO}}~\citep{li2023taco}.
CodeFeedback-Filtered-Instruction is a carefully curated dataset of code instruction queries, derived from four prominent open-source code instruction tuning datasets (e.g., Magicoder-OSS-Instruct).
TACO focuses on algorithmic code generation, featuring programming challenges from platforms such as LeetCode and Codeforces.
To prepare these datasets for unit test generation, we first apply a principle-driven approach to filter out unsuitable questions for unit testing with Llama3.1-70B (e.g., tasks involving randomness), as presented in Appendix~\ref{app:prompt}.
Subsequently, we restructure the original code solutions into functional formats to facilitate the following unit test generation.

\paragraph{Unit Test Generation.}
Based on the queries and code solutions in the dataset, we employ Llama3.1-70B to generate diverse unit tests via repeated sampling.
Each generated unit test is then executed on the ground truth code solution provided in the dataset.
The execution results serve as true/false labels for the unit tests, with correct unit tests enabling the code solution to successfully pass the test.
We observe that generating correct unit tests for difficult code instructions requires significant computational resources, often yielding only a small number of valid tests.
To address this, we utilize execution feedback from the Python interpreter to repair incorrect unit tests, rather than relying solely on repeated sampling.
By leveraging this feedback, Llama3.1-70B can efficiently repair tests, significantly improving the collection process for challenging problems.
To further identify high-quality unit tests, we employ a \textit{quality control} process.
We adhere to the principle that a high-quality unit test should allow the correct solution to pass while rejecting as many incorrect solutions as possible.
To achieve this, we generate incorrect solutions using a less capable model and filter out false positive unit tests (i.e., tests that fail to reject incorrect solutions).

\paragraph{Model Training}
We utilize supervised fine-tuning (SFT) to train \model based on Llama3.1-8B.
For the construction of training data, we use the problem and code solution as the instruction and the high-quality unit test as the answer.
An example of the instruction-answer pair is presented in Figure~\ref{fig:data_example}.

\subsection{Dynamic Unit Test Scaling}

Section~\ref{sec:scaling} presents that scaling unit tests is more effective for harder problems.
To further enhance the efficiency of unit test scaling, we implement a dynamic scaling strategy that adapts to problems of different difficulty, which is shown in the fourth section of Figure~\ref{fig:main}.
Specifically, we follow \citet{damani2024learning} to first train a problem difficulty classifier.
Subsequently, we use a greedy algorithm to allocate more computation resources on more challenging questions.

\paragraph{Problem Difficulty Estimation.}
To train the problem difficulty classifier, we first gather data using the preprocessed datasets from the synthetic data pipeline.
The policy model generates multiple solutions through repeated sampling, which are then evaluated using previously collected correct unit tests as verifiers.
The problem's difficulty is estimated by calculating the mean success probability (i.e., pass rate) of these solutions.

\paragraph{Model Training.}
We leverage the language model probing method~\citep{Alain2017porb, gurnee2024language} to train a lightweight problem difficulty classifier.
The probing method extracts implicit information from the intermediate representation, which is mapped into discrete classes by training a simple classifier~\citep{zhang-etal-2024-transferable, damani2024learning}.
Our problem difficulty classifier is a two-layer feedforward network with a scalar output to indicate the problem difficulties.
This classifier generates minimal overhead since its inputs are hidden states already calculated during the decoding process.
To train this classifier, we minimize the following cross-entropy loss:
\[
\sum_{x_i, \lambda_i} \!\left[ \lambda_i \log \!\left( \!\hat{\lambda}(x_i; \theta) \!\right) 
\!+ \!(1 \!- \!\lambda_i) \log \!\left( \!1 \!- \!\hat{\lambda}(x_i; \theta) \!\right) \!\right]
\]
where $x_i$ is the $i$-th query, $\theta$ is the parameters of the classifier, $\lambda_i$ is the actual pass rate, and $\hat{\lambda}(x_i; \theta)$ is the predicted pass rate (i.e., problem difficulty).


\paragraph{Dynamic Compution Allocation.}
We allocate more computation budgets on more challenging problems, following the method used by \citet{damani2024learning}.
For a question $x$ with pass rate $\lambda$,  the reward for allocating $b$ computation budget to this question is given by:
\begin{equation}
q(x, b) = 1 - (1 - \lambda)^b,
\label{eq:dynamic}
\end{equation}
which represents the probability of getting at least one correct answer in $b$ attempts.
Since for any question $x$, the reward function $q(x, b)$ is monotonically increasing with respect to $b$, we adopt a greedy algorithm to allocate computational resources across different questions.


\section{Experiments}
\label{sec:exp}
\subsection{Experimental Setup}

\paragraph{Datasets and Metrics.}
We conduct extensive experiments on three widely used benchmarks, including HumanEval Plus~\citep{liu2023evalplus}, MBPP Plus~\citep{liu2023evalplus}, and LiveCodeBench~\citep{jain2024livecodebench}.
HumanEval Plus and MBPP Plus are derived from HumanEval~\citep{chen2021humaneval} and MBPP~\citep{austin2021mbpp} with additional test cases.
LiveCodeBench is a contamination-free dataset that continuously collects new problems over time.
We select the queries from 2024-01 to 2024-09 to avoid the contamination problem with our training data.
We focus on the queries with solutions in functional format, ultimately yielding $168$ queries.
The evaluation metric is Pass$@k$~\citep{chen2021humaneval}, with $k$ set to $1$ in our experiments.

\begin{table}[t]
    \centering
    \setlength{\tabcolsep}{1.5pt}
    \scalebox{0.769}{
    \begin{tabular}{lllll}
    \toprule
    \multirow{2}{*}[-3pt]{Method} & \multicolumn{4}{c}{Policy Model} \\
    \cmidrule(lr){2-5}
    & Llama3-8B & Llama3-70B & GPT-3.5 & GPT-4o-m \\
    \midrule
    \grayline \multicolumn{5}{c}{\textbf{HumanEval Plus}} \\
    \midrule
    Vanilla & 53.58 & 73.74 & 67.83 & 82.96 \\
    \midrule
    Grading RM & 62.20\textsubscript{ +8.62} & 75.00\textsubscript{ +1.26} & 70.12\textsubscript{ +2.29} & 83.50\textsubscript{ +0.54} \\
    MBR-Exec & 60.30\textsubscript{ +6.72} & 75.80\textsubscript{ +2.06} & 70.60\textsubscript{ +2.77} & 85.20\textsubscript{ +2.24} \\
    CodeT & 65.30\textsubscript{ +11.72} & 76.20\textsubscript{ +2.46} & 73.89\textsubscript{ +6.06} & 85.30\textsubscript{ +2.34} \\
    MPSC & 59.72\textsubscript{ +6.14} & 75.51\textsubscript{ +1.77} & 72.76\textsubscript{ +4.93} & 84.82\textsubscript{ +1.86} \\
    \midrule
    Llama3.1-70B & \textbf{72.04}\textsubscript{ +18.46} & \underline{78.54}\textsubscript{ +4.80} & \textbf{79.76}\textsubscript{ +11.93} & \underline{85.45}\textsubscript{ +2.49} \\
    \model & \underline{72.01}\textsubscript{ +18.43} & \textbf{78.69}\textsubscript{ +4.95} & \underline{78.01}\textsubscript{ +10.18} & \textbf{86.38}\textsubscript{ +3.42} \\
    \midrule
    \grayline \multicolumn{5}{c}{\textbf{MBPP Plus}} \\
    \midrule
    Vanilla & 49.20 & 69.33 & 70.53 & 71.59 \\
    \midrule
    Grading RM & 48.40\textsubscript{ -0.80} & 70.60\textsubscript{ +1.27} & 66.67\textsubscript{ -3.86} & 69.00\textsubscript{ -2.59}\\
    MBR-Exec & 50.00\textsubscript{ +0.80} & 69.80\textsubscript{ +0.47} & 70.53\textsubscript{ +0.00} & 72.30\textsubscript{ +0.71}\\
    CodeT & 59.20\textsubscript{ +10.00} & 69.90\textsubscript{ +0.57} & 69.92\textsubscript{ -0.61} & 73.40\textsubscript{ +1.81} \\
    MPSC & 53.32\textsubscript{ +4.12} & 70.91\textsubscript{ +1.58} & 71.59\textsubscript{ +1.06} & 73.20\textsubscript{ +1.61} \\
    \midrule
    Llama3.1-70B & \underline{65.26}\textsubscript{ +16.06} & \underline{71.85}\textsubscript{ +2.52} & \underline{75.72}\textsubscript{ +5.19} & \underline{74.96}\textsubscript{ +3.37} \\
    \model & \textbf{66.71}\textsubscript{ +17.51} & \textbf{72.44}\textsubscript{ +3.11} & \textbf{75.96}\textsubscript{ +5.43} & \textbf{75.20}\textsubscript{ +3.61} \\
    \midrule
    \grayline \multicolumn{5}{c}{\textbf{LiveCodeBench}} \\
    \midrule
    Vanilla & 11.98 & 25.30 & 20.55 & 34.83 \\
    \midrule
    Grading RM & 13.10\textsubscript{ +1.12} & 26.19\textsubscript{ +0.89} & 20.83\textsubscript{ +0.28} & 36.31\textsubscript{ +1.48} \\
    MBR-Exec & 12.04\textsubscript{ +0.06} & 25.37\textsubscript{ +0.07} & 20.52\textsubscript{ -0.03} & 34.83\textsubscript{ +0.00} \\
    CodeT & 12.61\textsubscript{ +0.63} & 25.89\textsubscript{ +0.59} & 20.58\textsubscript{ +0.03} & 35.13\textsubscript{ +0.30} \\
    MPSC & 11.98\textsubscript{ +0.00} & 25.30\textsubscript{ +0.00} & 20.55\textsubscript{ +0.00} & 34.83\textsubscript{ +0.00} \\
    \midrule
    Llama3.1-70B & \underline{13.28}\textsubscript{ +1.30} & \textbf{28.46}\textsubscript{ +3.16} & \textbf{22.80}\textsubscript{ +2.25} & \underline{38.60}\textsubscript{ +3.77} \\
    \model & \textbf{15.21}\textsubscript{ +3.23} & \underline{27.73}\textsubscript{ +2.43} & \underline{21.76}\textsubscript{ +1.21} & \textbf{39.20}\textsubscript{ +4.37} \\
    \bottomrule
    \end{tabular}
    }
    \caption{
    The main result of our approach and other baselines over three code generation benchmarks.
    GPT-4o-m stands for GPT-4o-mini.
    The improvements are calculated between methods and vanilla.
    The top two performances for each dataset and policy model are marked in \textbf{bold} and \underline{underlined}.
    }
    \label{tab:main_result}
\end{table}

\paragraph{Baselines and Implementations.}
We compare several baselines on four policy models, including Llama3-8B~\citep{llama3-report}, Llama3-70B~\citep{llama3-report}, GPT-3.5-turbo\footnote{\url{https://chat.openai.com/}}, and GPT-4o-mini~\cite{gpt-4-report}.
The baselines include the vanilla method that randomly selects a solution, the grading reward model, MBR-Exec~\citep{shi2022mbr-exec}, CodeT~\citep{chen2023codet}, and MPSC~\citep{huang-etal-2024-mpsc}.
For the grading reward model, we employ ArmoRM-Llama3-8B-v0.1~\citep{ArmoRM} and generate a scalar score for each candidate solution.
For other baselines, we utilize Llama3.1-70B to generate test cases, using the same computation budget to conduct fair comparisons. 
For our method, we employ the majority voting framework in Section~\ref{sec:methodology} and utilize Llama3.1-70B and \model as unit test generator.
Appendix~\ref{app:baseline} presents the detailed experimental setting and the implementation of all baselines.

\begin{figure}[t]
  \centering
  \includegraphics[width=0.47\textwidth]{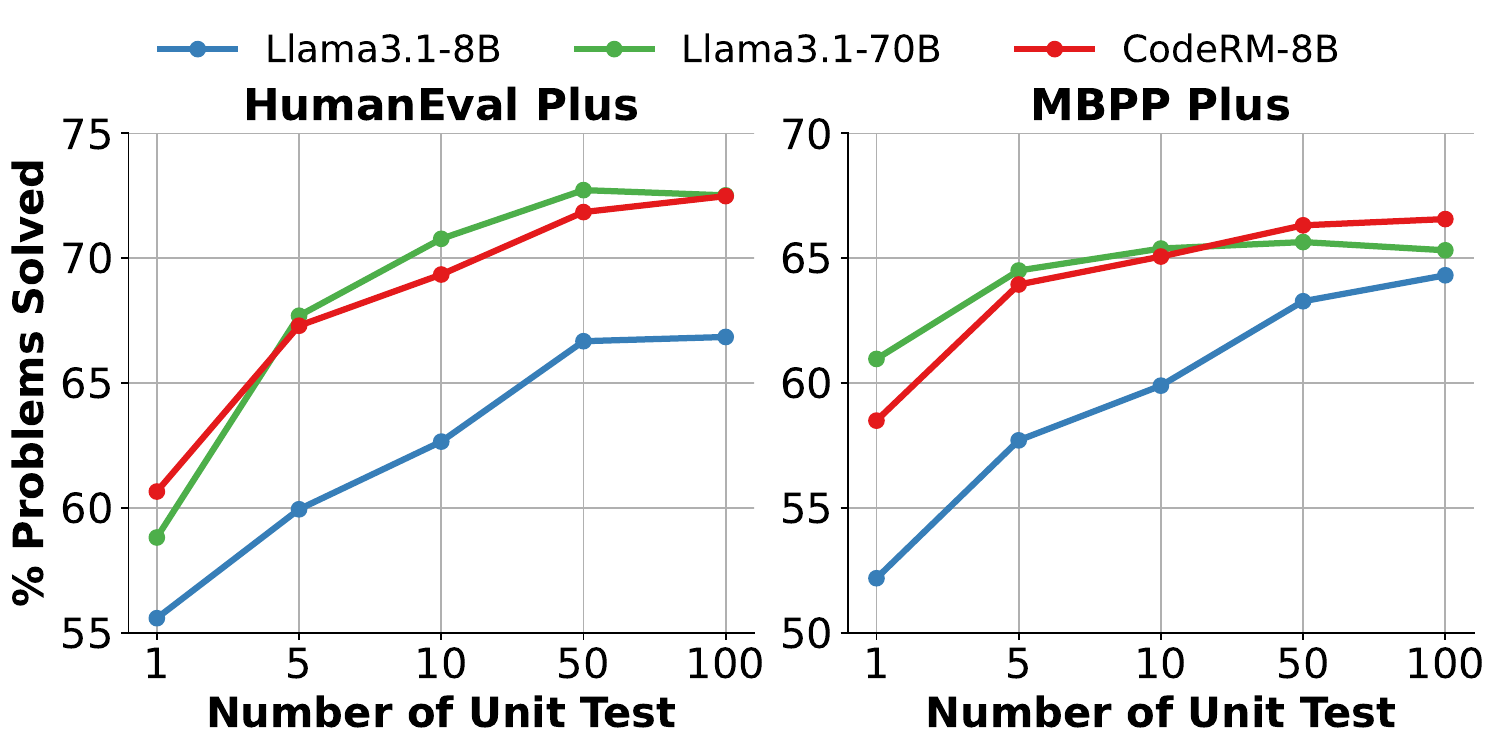}
  \caption{The performance of three different unit test generators (reward model) on different quantities of unit tests, while employing Llama3-8B as the policy model.}
 \label{fig:result_on_scale}
\end{figure}

\begin{figure*}[t]
  \centering
  \includegraphics[width=0.9\textwidth]{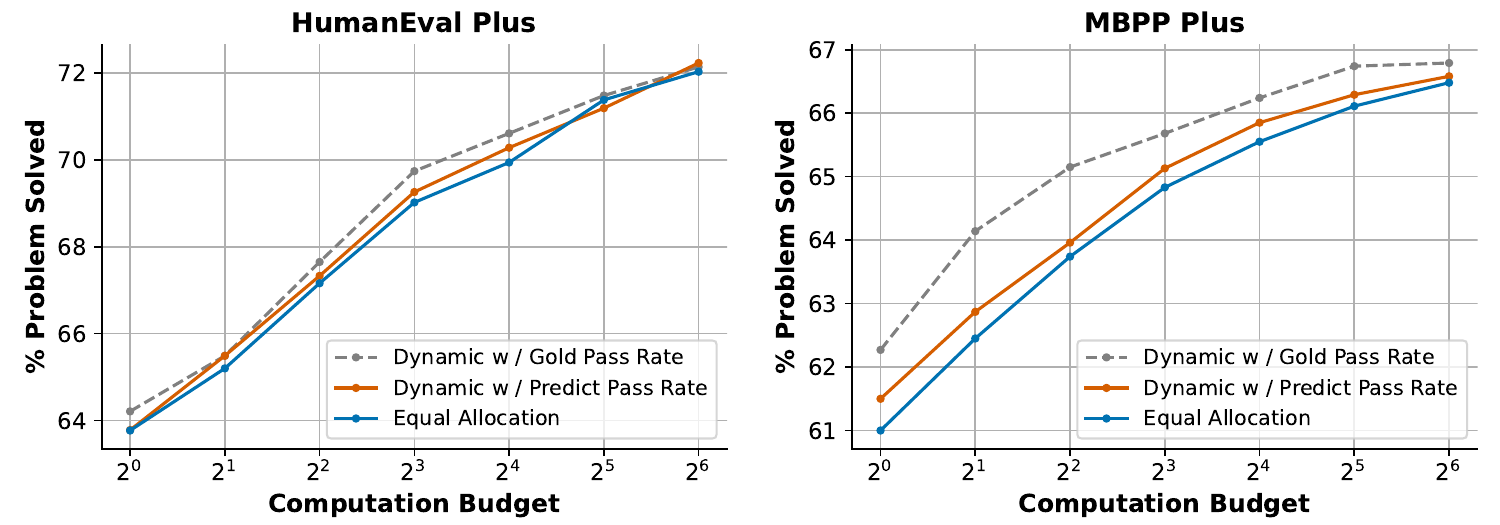}
  \caption{Best-of-N performance comparison under unit test scaling with three computation budget allocation strategies: dynamic allocation with gold pass rate, dynamic allocation with predicted pass rate, and equal allocation.}
 \label{fig:dynamic_scale}
\end{figure*}

\subsection{Main Results}

\paragraph{Effectiveness of \model.}
Table~\ref{tab:main_result} summarizes the experimental results on three code generation benchmarks.
Under our unit test-based majority voting framework, both Llama3.1-70B and \model significantly improve the performance of four policy models across all benchmarks.
Notably, the best-of-N performance of \model is on par with Llama3.1-70B, a model with nearly 8x more parameters.
Across all three benchmarks, \model demonstrates consistent and substantial enhancements to the performance of policy models at varying scales and types.
For example, on HumanEval Plus, \model achieves a substantial performance improvement of 18.43\% for the smaller model Llama3-8B, while also enhancing the performance of larger models and proprietary models, such as Llama3-70B and GPT-4o-mini, by 4.95\% and 3.42\%, respectively.
Figure~\ref{fig:result_on_scale} visualizes the performance of various unit test generators as the number of unit tests scales.
The results highlight that \model achieves performance comparable to Llama3.1-70B while substantially surpassing Llama3.1-8B, emphasizing its effectiveness and computational efficiency.


\paragraph{Performance of Dynamic Scaling.}
Figure~\ref{fig:dynamic_scale} presents the performance of our dynamic unit test scaling implementation, which leverages Equation~\eqref{eq:dynamic} to allocate computation budgets based on predicted pass rates.
We compare it against two baselines: 1) \textit{dynamic allocation with gold pass rates}, which uses actual pass rates instead of predictions, and 2) \textit{equal allocation}, which does not apply dynamic scaling.
Results show that our method improves performance under fixed computational budgets, with more pronounced gains on MBPP Plus due to its higher question difficulty compared to HumanEval Plus, making dynamic scaling more impactful.
Future work could explore more accurate difficulty classifiers or alternative allocation algorithms to further improve performance.

\subsection{Quality of Generated Unit Tests}
\label{sec:ut_quality}

\begin{table}[t]
  \setlength{\tabcolsep}{4pt}
  \begin{center}
    \scalebox{0.83}{
    \begin{tabular} {lcccc}
      \toprule
      Model & Acc ($\uparrow$) & F1 ($\uparrow$) & FAR ($\downarrow$) & FRR ($\downarrow$) \\
      \midrule
      \grayline \multicolumn{5}{c}{Quality of Individual Unit Tests} \\
      \midrule
      Llama3.1-8B & 60.02 & 44.97 & 13.66 & 46.13 \\
      Llama3.1-70B & \textbf{73.65} & \textbf{70.15} & \textbf{11.10} & \textbf{34.51} \\
      \model (Ours) & \underline{69.64} & \underline{63.63} & \underline{11.17} & \underline{38.55} \\
      \midrule
      \grayline \multicolumn{5}{c}{Quality of Multiple Unit Tests} \\
      \midrule
      Llama3.1-8B & 74.21 & 74.35 & 20.44 & 30.55 \\
      Llama3.1-70B & \underline{78.30} & \underline{78.76} & \underline{17.19} & \underline{25.97} \\
      \model (Ours) & \textbf{80.46} & \textbf{81.27} & \textbf{16.48} & \textbf{22.71} \\
      \bottomrule
    \end{tabular}
    }
    \caption{The quality of individual unit tests and the combination of multiple unit tests on HumanEval Plus, utilizing Llama3.1-8B as the policy model.
    The top two performances are highlighted using \textbf{bold} and \underline{underlining}.
    }
    \label{tab:ut_quality}
  \end{center}
\end{table}

We evaluate the quality of the unit test generated by \model.
As each unit test functions as a classifier to determine correct or incorrect solutions, we first utilize accuracy and F1 score as metrics to assess the classification performance of the unit test.
We further propose two new metrics to detailed evaluate the possibility of the unit test making incorrect judgments.
False Acceptance Rate (FAR) measures the probability that unit tests incorrectly accept invalid solutions.
False Rejection Rate (FRR) measures the probability that unit tests incorrectly reject valid solutions.
The calculation formulas for these four metrics are introduced in Appendix~\ref{app:metrics}.

Table~\ref{tab:ut_quality} presents the quality of reward signal produced by an individual unit test and the combination of $100$ unit tests under our majority voting framework.
The results indicate that the trained \model renders more precise assessments of solutions and makes fewer errors than the original Llama3.1-8B.
Moreover, we note that while the quality of individual unit tests of \model is inferior to Llama3.1-70B, the quality of multiple unit tests is superior.
This phenomenon may suggest that \model produces more diverse unit tests, offering diverse perspectives within our majority voting framework and resulting in a higher quality of code reward signal.

\subsection{Ablation Study of Synthetic Data}

We conduct ablation studies to investigate the effectiveness of data quality control and instruction-tuning data size in our synthetic data pipeline.

\begin{table}[t]
  \setlength{\tabcolsep}{4pt}
  \begin{center}
    \scalebox{0.83}{
    \begin{tabular} {lll}
      \toprule
      Method & HumanEval+ & MBPP+  \\
      \midrule
      zero-shot & 66.67 & 63.27 \\
      training wo / quality control & 69.71\textsubscript{ +3.04} & 64.96\textsubscript{ +1.69} \\
      training w / quality control & 71.09\textsubscript{ +4.42} & 66.31\textsubscript{ +3.04} \\
      \bottomrule
    \end{tabular}
    }
    \caption{The effects of synthetic data quality control.}
    \label{tab:abl_quality}
  \end{center}
\end{table}

\paragraph{Data Quality.}
Filtering false positive unit tests is a critical component for increasing training data quality in our synthetic data pipeline.
We evaluate the performance of \model against the trained model that lacks this quality control procedure.
Table~\ref{tab:abl_quality} presents that the quality control procedure significantly increased the performance of the trained model,  with relative performance gains of approximately $45\%$ and $80\%$ on HumanEval Plus and MBPP Plus, respectively, compared to training without quality control.

\begin{figure}[t]
  \centering
  \includegraphics[width=0.47\textwidth]{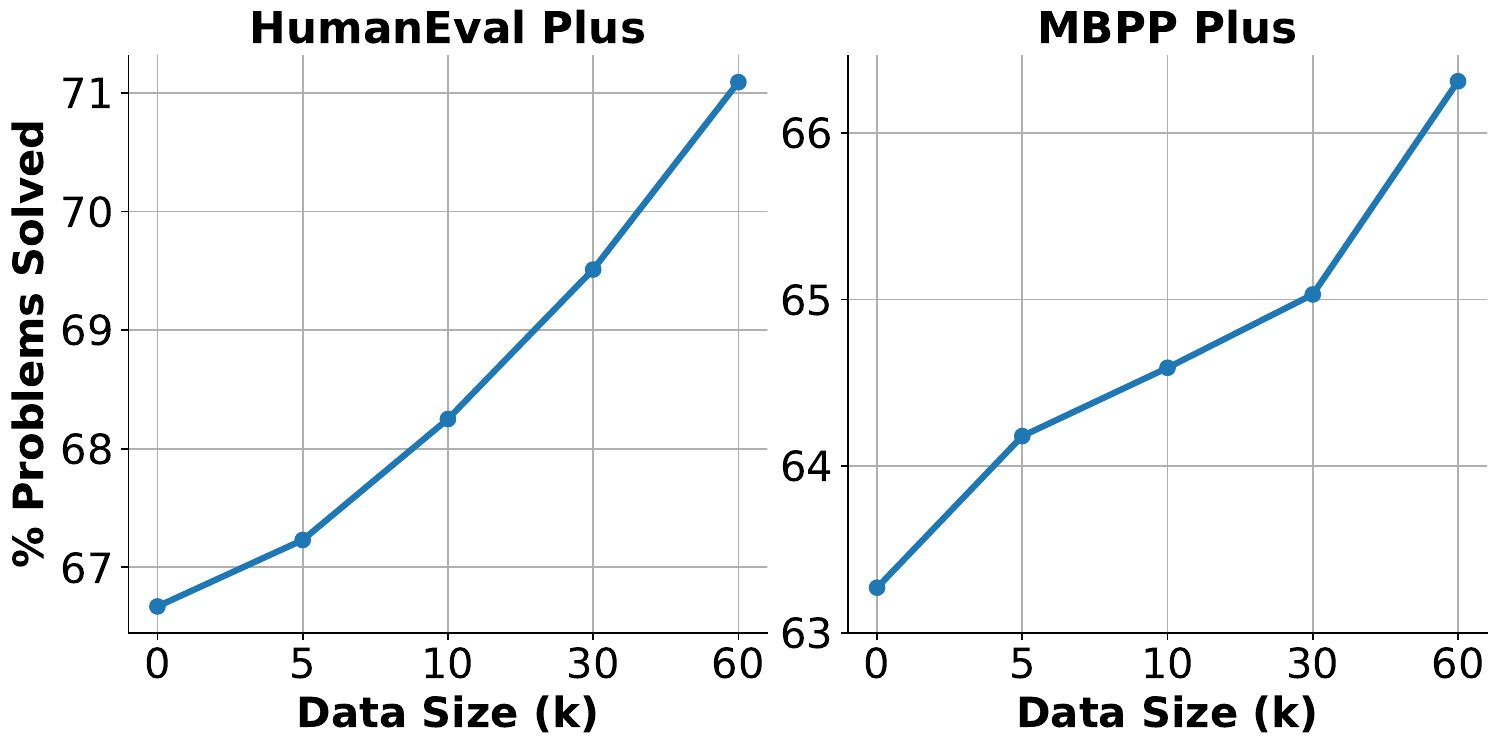}
  \caption{The effects of data size.}
 \label{fig:abl_on_data_size}
\end{figure}

\paragraph{Data Size.}
Data size is also a crucial factor in enhancing the model's performance.
We evaluate the performance of the model trained on different amounts of training data.
Figure~\ref{fig:abl_on_data_size} presents that as the data size increases, the model's performance consistently improves.
This observation demonstrates that collecting more high-quality instruction-tuning data and using our data synthesis pipeline to generate more high-quality unit tests can substantially enhance model performance.

\section{Related Work}
\label{sec:related_work}
\paragraph{Reranking and selection for optimal solutions.}
Current reranking techniques for optimal solution selection can be classified into two categories: execution-based and non-execution-based methods.
Execution-based methods leverage test cases as verifiers for selecting optimal code solutions~\citep{li2022alphacode, shi2022mbr-exec, chen2023codet, zhang2023algo, chen2024b4}.
For example, \citet{shi2022mbr-exec} employ execution result-based minimum Bayes risk decoding for solution selection.
\citet{chen2023codet} evaluate output consistency with generated test cases and concordance with other candidate code solutions.
Non-execution-based methods use deep learning-based rerankers for optimal code solution selection~\citep{inala2022faultaware, zhang2023codereviewer}.
\citet{inala2022faultaware} trains a neural reranker utilizing the code execution results as labels for classification.
\citet{zhang2023codereviewer} utilizes a collaborative system that employs additional reviewer models to assess the likelihood of the instructions based on the generated code solutions.
In this paper, we discover the effects of dynamic scaling of unit tests over programming problems.

\paragraph{Unit test generation.}
The expense of human-created and maintained unit tests prompts the advancement of automatic unit test generation techniques.
Traditional methods, such as search-based~\citep{harman2010testing, pynguin}, constraint-based~\citep{6693084}, and probability-based approaches~\citep{4222570}, often achieve acceptable correctness but suffer from limited coverage, poor readability, and are typically restricted to regression or implicit oracles~\citep{oracle_survey}.
Recently, deep learning models, particularly large language models, have gained traction for unit test generation~\citep{a3test, ChatUniTest, ut_survey, ut_gpt}.
In this paper, we propose a data synthetic pipeline and train a model for high-quality unit test generation.


\section{Conclusion}
This paper investigates the impact of scaling unit tests on improving the quality of code reward signals.
Pioneer results demonstrate a positive correlation between unit test number and reward signal quality, with more greater benefits observed in challenging problems.
To facilitate efficient and high-quality unit test scaling, we train a small yet powerful unit test generator and implement a dynamic scaling strategy.
Experimental results demonstrate that our approach significantly boosts the performance of models in various parameter sizes.




\section{Limitations}
\paragraph{Implementation of Dynamic Scaling.}
Our implementation of dynamic unit test scaling is based on the method proposed by \citet{damani2024learning}, which allocates more computational resources to harder problems.
However, while \citet{damani2024learning} directly optimizes the allocation of computational resources for the policy model (i.e., models that generate responses), their approach does not directly extend to optimizing the reward model (i.e., models that generate verifiers).
Consequently, directly adopting this method may not be entirely appropriate in our context.
As shown in Figure~\ref{fig:dynamic_scale}, the performance improvement is relatively modest when computation budgets are allocated based on the gold pass rate on HumanEval Plus.
Future research could explore more effective methods for dynamically scaling the number of unit tests based on varying problem difficulties.

\paragraph{Diversity and Coverage of Unit Test.}
We observe in Section~\ref{sec:scaling} that Gemma-2-27B-it performs significantly worse than Llama3.1-70B when using a single unit test.
However, its performance becomes comparable when scaled to 100 unit tests per question.
This observation suggests that Gemma-2-27B-it may generate more diverse and higher-coverage unit tests than Llama3.1-70B through repeated sampling.
Future work could explore the diversity and coverage of unit tests as the number of unit tests scales, which may provide insights into training more effective reward models for improved supervision accuracy.

\bibliography{custom}

\begin{thebibliography}{44}
\providecommand{\natexlab}[1]{#1}

\bibitem[{Achiam et~al.(2023)Achiam, Adler, Agarwal, Ahmad, Akkaya, Aleman, Almeida, Altenschmidt, Altman, Anadkat et~al.}]{gpt-4-report}
Josh Achiam, Steven Adler, Sandhini Agarwal, Lama Ahmad, Ilge Akkaya, Florencia~Leoni Aleman, Diogo Almeida, Janko Altenschmidt, Sam Altman, Shyamal Anadkat, et~al. 2023.
\newblock \href {https://arxiv.org/abs/2303.08774} {Gpt-4 technical report}.
\newblock \emph{arXiv preprint arXiv:2303.08774}.

\bibitem[{Alagarsamy et~al.(2024)Alagarsamy, Tantithamthavorn, and Aleti}]{a3test}
Saranya Alagarsamy, Chakkrit Tantithamthavorn, and Aldeida Aleti. 2024.
\newblock \href {https://doi.org/10.1016/j.infsof.2024.107565} {A3test: Assertion-augmented automated test case generation}.
\newblock \emph{Inf. Softw. Technol.}, 176(C).

\bibitem[{Alain and Bengio(2017)}]{Alain2017porb}
Guillaume Alain and Yoshua Bengio. 2017.
\newblock \href {https://openreview.net/forum?id=HJ4-rAVtl} {Understanding intermediate layers using linear classifier probes}.
\newblock In \emph{ICLR (Workshop)}.

\bibitem[{Austin et~al.(2021)Austin, Odena, Nye, Bosma, Michalewski, Dohan, Jiang, Cai, Terry, Le et~al.}]{austin2021mbpp}
Jacob Austin, Augustus Odena, Maxwell Nye, Maarten Bosma, Henryk Michalewski, David Dohan, Ellen Jiang, Carrie Cai, Michael Terry, Quoc Le, et~al. 2021.
\newblock \href {https://arxiv.org/abs/2108.07732} {Program synthesis with large language models}.
\newblock \emph{arXiv preprint arXiv:2108.07732}.

\bibitem[{Bansal et~al.(2024)Bansal, Hosseini, Agarwal, Tran, and Kazemi}]{bansal2024smaller}
Hritik Bansal, Arian Hosseini, Rishabh Agarwal, Vinh~Q Tran, and Mehran Kazemi. 2024.
\newblock \href {https://arxiv.org/abs/2408.16737} {Smaller, weaker, yet better: Training llm reasoners via compute-optimal sampling}.
\newblock \emph{arXiv preprint arXiv:2408.16737}.

\bibitem[{Barr et~al.(2015)Barr, Harman, McMinn, Shahbaz, and Yoo}]{oracle_survey}
Earl~T. Barr, Mark Harman, Phil McMinn, Muzammil Shahbaz, and Shin Yoo. 2015.
\newblock \href {https://doi.org/10.1109/TSE.2014.2372785} {The oracle problem in software testing: A survey}.
\newblock \emph{IEEE Transactions on Software Engineering}, 41(5):507--525.

\bibitem[{Brown et~al.(2024)Brown, Juravsky, Ehrlich, Clark, Le, R{\'e}, and Mirhoseini}]{brown2024monkey}
Bradley Brown, Jordan Juravsky, Ryan Ehrlich, Ronald Clark, Quoc~V Le, Christopher R{\'e}, and Azalia Mirhoseini. 2024.
\newblock \href {https://arxiv.org/abs/2407.21787} {Large language monkeys: Scaling inference compute with repeated sampling}.
\newblock \emph{arXiv preprint arXiv:2407.21787}.

\bibitem[{Chen et~al.(2023)Chen, Zhang, Nguyen, Zan, Lin, Lou, and Chen}]{chen2023codet}
Bei Chen, Fengji Zhang, Anh Nguyen, Daoguang Zan, Zeqi Lin, Jian-Guang Lou, and Weizhu Chen. 2023.
\newblock \href {https://openreview.net/forum?id=ktrw68Cmu9c} {Codet: Code generation with generated tests}.
\newblock In \emph{The Eleventh International Conference on Learning Representations}.

\bibitem[{Chen et~al.(2021)Chen, Tworek, Jun, Yuan, Pinto, Kaplan, Edwards, Burda, Joseph, Brockman et~al.}]{chen2021humaneval}
Mark Chen, Jerry Tworek, Heewoo Jun, Qiming Yuan, Henrique Ponde De~Oliveira Pinto, Jared Kaplan, Harri Edwards, Yuri Burda, Nicholas Joseph, Greg Brockman, et~al. 2021.
\newblock \href {https://arxiv.org/abs/2107.03374} {Evaluating large language models trained on code}.
\newblock \emph{arXiv preprint arXiv:2107.03374}.

\bibitem[{Chen et~al.(2024{\natexlab{a}})Chen, Liu, Tao, Hong, Lo, Xia, and Sun}]{chen2024b4}
Mouxiang Chen, Zhongxin Liu, He~Tao, Yusu Hong, David Lo, Xin Xia, and Jianling Sun. 2024{\natexlab{a}}.
\newblock \href {https://doi.org/10.1145/3691620.3695536} {B4: Towards optimal assessment of plausible code solutions with plausible tests}.
\newblock In \emph{Proceedings of the 39th IEEE/ACM International Conference on Automated Software Engineering}, ASE '24, page 1693–1705, New York, NY, USA. Association for Computing Machinery.

\bibitem[{Chen et~al.(2024{\natexlab{b}})Chen, Hu, Zhi, Han, Deng, and Yin}]{ChatUniTest}
Yinghao Chen, Zehao Hu, Chen Zhi, Junxiao Han, Shuiguang Deng, and Jianwei Yin. 2024{\natexlab{b}}.
\newblock \href {https://doi.org/10.1145/3663529.3663801} {Chatunitest: A framework for llm-based test generation}.
\newblock In \emph{Companion Proceedings of the 32nd ACM International Conference on the Foundations of Software Engineering}, FSE 2024, page 572–576, New York, NY, USA. Association for Computing Machinery.

\bibitem[{Cobbe et~al.(2021)Cobbe, Kosaraju, Bavarian, Chen, Jun, Kaiser, Plappert, Tworek, Hilton, Nakano et~al.}]{cobbe2021training}
Karl Cobbe, Vineet Kosaraju, Mohammad Bavarian, Mark Chen, Heewoo Jun, Lukasz Kaiser, Matthias Plappert, Jerry Tworek, Jacob Hilton, Reiichiro Nakano, et~al. 2021.
\newblock \href {https://arxiv.org/abs/2110.14168} {Training verifiers to solve math word problems}.
\newblock \emph{arXiv preprint arXiv:2110.14168}.

\bibitem[{Damani et~al.(2024)Damani, Shenfeld, Peng, Bobu, and Andreas}]{damani2024learning}
Mehul Damani, Idan Shenfeld, Andi Peng, Andreea Bobu, and Jacob Andreas. 2024.
\newblock \href {https://arxiv.org/abs/2410.04707} {Learning how hard to think: Input-adaptive allocation of lm computation}.
\newblock \emph{arXiv preprint arXiv:2410.04707}.

\bibitem[{Dubey et~al.(2024)Dubey, Jauhri, Pandey, Kadian, Al-Dahle, Letman, Mathur, Schelten, Yang, Fan et~al.}]{llama3-report}
Abhimanyu Dubey, Abhinav Jauhri, Abhinav Pandey, Abhishek Kadian, Ahmad Al-Dahle, Aiesha Letman, Akhil Mathur, Alan Schelten, Amy Yang, Angela Fan, et~al. 2024.
\newblock \href {https://arxiv.org/abs/2407.21783} {The llama 3 herd of models}.
\newblock \emph{arXiv preprint arXiv:2407.21783}.

\bibitem[{Edmonds(1971)}]{edmonds1971matroids}
Jack Edmonds. 1971.
\newblock \href {https://doi.org/10.1007/BF01584082} {Matroids and the greedy algorithm}.
\newblock \emph{Mathematical programming}, 1:127--136.

\bibitem[{Efron(1979)}]{1976bootstrap}
B.~Efron. 1979.
\newblock \href {https://doi.org/10.1214/aos/1176344552} {{Bootstrap Methods: Another Look at the Jackknife}}.
\newblock \emph{The Annals of Statistics}, 7(1):1 -- 26.

\bibitem[{Gao et~al.(2023)Gao, Schulman, and Hilton}]{gao2023scalingrm}
Leo Gao, John Schulman, and Jacob Hilton. 2023.
\newblock \href {https://proceedings.mlr.press/v202/gao23h.html} {Scaling laws for reward model overoptimization}.
\newblock In \emph{International Conference on Machine Learning}, pages 10835--10866. PMLR.

\bibitem[{Gurnee and Tegmark(2024)}]{gurnee2024language}
Wes Gurnee and Max Tegmark. 2024.
\newblock \href {https://openreview.net/forum?id=jE8xbmvFin} {Language models represent space and time}.
\newblock In \emph{The Twelfth International Conference on Learning Representations}.

\bibitem[{Harman and McMinn(2010)}]{harman2010testing}
Mark Harman and Phil McMinn. 2010.
\newblock \href {https://doi.org/10.1109/TSE.2009.71} {A theoretical and empirical study of search-based testing: Local, global, and hybrid search}.
\newblock \emph{IEEE Transactions on Software Engineering}, 36(2):226--247.

\bibitem[{Huang et~al.(2024{\natexlab{a}})Huang, Lu, Wan, and Duan}]{huang-etal-2024-mpsc}
Baizhou Huang, Shuai Lu, Xiaojun Wan, and Nan Duan. 2024{\natexlab{a}}.
\newblock \href {https://doi.org/10.18653/v1/2024.acl-long.78} {Enhancing large language models in coding through multi-perspective self-consistency}.
\newblock In \emph{Proceedings of the 62nd Annual Meeting of the Association for Computational Linguistics (Volume 1: Long Papers)}, pages 1429--1450, Bangkok, Thailand. Association for Computational Linguistics.

\bibitem[{Huang et~al.(2024{\natexlab{b}})Huang, Chen, Mishra, Zheng, Yu, Song, and Zhou}]{huang2024large}
Jie Huang, Xinyun Chen, Swaroop Mishra, Huaixiu~Steven Zheng, Adams~Wei Yu, Xinying Song, and Denny Zhou. 2024{\natexlab{b}}.
\newblock \href {https://openreview.net/forum?id=IkmD3fKBPQ} {Large language models cannot self-correct reasoning yet}.
\newblock In \emph{The Twelfth International Conference on Learning Representations}.

\bibitem[{Huang et~al.(2024{\natexlab{c}})Huang, Yu, Ma, Zhong, Feng, Wang, Chen, Peng, Feng, Qin, and Liu}]{huang2024hallu-survey}
Lei Huang, Weijiang Yu, Weitao Ma, Weihong Zhong, Zhangyin Feng, Haotian Wang, Qianglong Chen, Weihua Peng, Xiaocheng Feng, Bing Qin, and Ting Liu. 2024{\natexlab{c}}.
\newblock \href {https://doi.org/10.1145/3703155} {A survey on hallucination in large language models: Principles, taxonomy, challenges, and open questions}.
\newblock \emph{ACM Transactions on Information Systems}.

\bibitem[{Inala et~al.(2022)Inala, Wang, Yang, Codas, Encarnaci{\'o}n, Lahiri, Musuvathi, and Gao}]{inala2022faultaware}
Jeevana~Priya Inala, Chenglong Wang, Mei Yang, Andres Codas, Mark Encarnaci{\'o}n, Shuvendu~K Lahiri, Madanlal Musuvathi, and Jianfeng Gao. 2022.
\newblock \href {https://openreview.net/forum?id=LtJMqnbslJe} {Fault-aware neural code rankers}.
\newblock In \emph{Advances in Neural Information Processing Systems}.

\bibitem[{Jain et~al.(2024)Jain, Han, Gu, Li, Yan, Zhang, Wang, Solar-Lezama, Sen, and Stoica}]{jain2024livecodebench}
Naman Jain, King Han, Alex Gu, Wen-Ding Li, Fanjia Yan, Tianjun Zhang, Sida Wang, Armando Solar-Lezama, Koushik Sen, and Ion Stoica. 2024.
\newblock \href {https://arxiv.org/abs/2403.07974} {Livecodebench: Holistic and contamination free evaluation of large language models for code}.
\newblock \emph{arXiv preprint arXiv:2403.07974}.

\bibitem[{Kadavath et~al.(2022)Kadavath, Conerly, Askell, Henighan, Drain, Perez, Schiefer, Hatfield-Dodds, DasSarma, Tran-Johnson et~al.}]{kadavath2022language}
Saurav Kadavath, Tom Conerly, Amanda Askell, Tom Henighan, Dawn Drain, Ethan Perez, Nicholas Schiefer, Zac Hatfield-Dodds, Nova DasSarma, Eli Tran-Johnson, et~al. 2022.
\newblock \href {https://arxiv.org/abs/2207.05221} {Language models (mostly) know what they know}.
\newblock \emph{arXiv preprint arXiv:2207.05221}.

\bibitem[{Li et~al.(2023)Li, Fu, Zhang, Huang, Sun, Lyu, Liu, Jin, and Li}]{li2023taco}
Rongao Li, Jie Fu, Bo-Wen Zhang, Tao Huang, Zhihong Sun, Chen Lyu, Guang Liu, Zhi Jin, and Ge~Li. 2023.
\newblock \href {https://arxiv.org/abs/2312.14852} {Taco: Topics in algorithmic code generation dataset}.
\newblock \emph{arXiv preprint arXiv:2312.14852}.

\bibitem[{Li et~al.(2022)Li, Choi, Chung, Kushman, Schrittwieser, Leblond, Eccles, Keeling, Gimeno, Lago, Hubert, Choy, de~Masson~d’Autume, Babuschkin, Chen, Huang, Welbl, Gowal, Cherepanov, Molloy, Mankowitz, Robson, Kohli, de~Freitas, Kavukcuoglu, and Vinyals}]{li2022alphacode}
Yujia Li, David Choi, Junyoung Chung, Nate Kushman, Julian Schrittwieser, Rémi Leblond, Tom Eccles, James Keeling, Felix Gimeno, Agustin~Dal Lago, Thomas Hubert, Peter Choy, Cyprien de~Masson~d’Autume, Igor Babuschkin, Xinyun Chen, Po-Sen Huang, Johannes Welbl, Sven Gowal, Alexey Cherepanov, James Molloy, Daniel~J. Mankowitz, Esme~Sutherland Robson, Pushmeet Kohli, Nando de~Freitas, Koray Kavukcuoglu, and Oriol Vinyals. 2022.
\newblock \href {https://doi.org/10.1126/science.abq1158} {Competition-level code generation with alphacode}.
\newblock \emph{Science}, 378(6624):1092--1097.

\bibitem[{Lightman et~al.(2024)Lightman, Kosaraju, Burda, Edwards, Baker, Lee, Leike, Schulman, Sutskever, and Cobbe}]{lightman2023let}
Hunter Lightman, Vineet Kosaraju, Yuri Burda, Harrison Edwards, Bowen Baker, Teddy Lee, Jan Leike, John Schulman, Ilya Sutskever, and Karl Cobbe. 2024.
\newblock \href {https://openreview.net/forum?id=v8L0pN6EOi} {Let's verify step by step}.
\newblock In \emph{The Twelfth International Conference on Learning Representations}.

\bibitem[{Liu et~al.(2023)Liu, Xia, Wang, and ZHANG}]{liu2023evalplus}
Jiawei Liu, Chunqiu~Steven Xia, Yuyao Wang, and LINGMING ZHANG. 2023.
\newblock \href {https://openreview.net/forum?id=1qvx610Cu7} {Is your code generated by chat{GPT} really correct? rigorous evaluation of large language models for code generation}.
\newblock In \emph{Thirty-seventh Conference on Neural Information Processing Systems}.

\bibitem[{Lukasczyk and Fraser(2022)}]{pynguin}
Stephan Lukasczyk and Gordon Fraser. 2022.
\newblock \href {https://doi.org/10.1145/3510454.3516829} {Pynguin: automated unit test generation for python}.
\newblock In \emph{Proceedings of the ACM/IEEE 44th International Conference on Software Engineering: Companion Proceedings}, ICSE '22, page 168–172, New York, NY, USA. Association for Computing Machinery.

\bibitem[{Pacheco et~al.(2007)Pacheco, Lahiri, Ernst, and Ball}]{4222570}
Carlos Pacheco, Shuvendu~K. Lahiri, Michael~D. Ernst, and Thomas Ball. 2007.
\newblock \href {https://doi.org/10.1109/ICSE.2007.37} {Feedback-directed random test generation}.
\newblock In \emph{29th International Conference on Software Engineering (ICSE'07)}, pages 75--84.

\bibitem[{Schäfer et~al.(2024)Schäfer, Nadi, Eghbali, and Tip}]{ut_survey}
Max Schäfer, Sarah Nadi, Aryaz Eghbali, and Frank Tip. 2024.
\newblock \href {https://doi.org/10.1109/TSE.2023.3334955} {An empirical evaluation of using large language models for automated unit test generation}.
\newblock \emph{IEEE Transactions on Software Engineering}, 50(1):85--105.

\bibitem[{Shi et~al.(2022)Shi, Fried, Ghazvininejad, Zettlemoyer, and Wang}]{shi2022mbr-exec}
Freda Shi, Daniel Fried, Marjan Ghazvininejad, Luke Zettlemoyer, and Sida~I. Wang. 2022.
\newblock \href {https://doi.org/10.18653/v1/2022.emnlp-main.231} {Natural language to code translation with execution}.
\newblock In \emph{Proceedings of the 2022 Conference on Empirical Methods in Natural Language Processing}, pages 3533--3546, Abu Dhabi, United Arab Emirates. Association for Computational Linguistics.

\bibitem[{Snell et~al.(2024)Snell, Lee, Xu, and Kumar}]{snell2024scaling}
Charlie Snell, Jaehoon Lee, Kelvin Xu, and Aviral Kumar. 2024.
\newblock \href {https://arxiv.org/abs/2408.03314} {Scaling llm test-time compute optimally can be more effective than scaling model parameters}.
\newblock \emph{arXiv preprint arXiv:2408.03314}.

\bibitem[{Svyatkovskiy et~al.(2020)Svyatkovskiy, Deng, Fu, and Sundaresan}]{codegen2020}
Alexey Svyatkovskiy, Shao~Kun Deng, Shengyu Fu, and Neel Sundaresan. 2020.
\newblock \href {https://doi.org/10.1145/3368089.3417058} {Intellicode compose: code generation using transformer}.
\newblock In \emph{Proceedings of the 28th ACM Joint Meeting on European Software Engineering Conference and Symposium on the Foundations of Software Engineering}, ESEC/FSE 2020, page 1433–1443, New York, NY, USA. Association for Computing Machinery.

\bibitem[{Touvron et~al.(2023)Touvron, Lavril, Izacard, Martinet, Lachaux, Lacroix, Rozi{\`e}re, Goyal, Hambro, Azhar et~al.}]{touvron2023llama}
Hugo Touvron, Thibaut Lavril, Gautier Izacard, Xavier Martinet, Marie-Anne Lachaux, Timoth{\'e}e Lacroix, Baptiste Rozi{\`e}re, Naman Goyal, Eric Hambro, Faisal Azhar, et~al. 2023.
\newblock \href {https://arxiv.org/abs/2302.13971} {Llama: Open and efficient foundation language models}.
\newblock \emph{arXiv preprint arXiv:2302.13971}.

\bibitem[{Wang et~al.(2024)Wang, Xiong, Xie, Zhao, and Zhang}]{ArmoRM}
Haoxiang Wang, Wei Xiong, Tengyang Xie, Han Zhao, and Tong Zhang. 2024.
\newblock \href {https://doi.org/10.18653/v1/2024.findings-emnlp.620} {Interpretable preferences via multi-objective reward modeling and mixture-of-experts}.
\newblock In \emph{Findings of the Association for Computational Linguistics: EMNLP 2024}, pages 10582--10592, Miami, Florida, USA. Association for Computational Linguistics.

\bibitem[{Wang et~al.(2023)Wang, Wei, Schuurmans, Le, Chi, Narang, Chowdhery, and Zhou}]{wang2023selfconsistency}
Xuezhi Wang, Jason Wei, Dale Schuurmans, Quoc~V Le, Ed~H. Chi, Sharan Narang, Aakanksha Chowdhery, and Denny Zhou. 2023.
\newblock \href {https://openreview.net/forum?id=1PL1NIMMrw} {Self-consistency improves chain of thought reasoning in language models}.
\newblock In \emph{The Eleventh International Conference on Learning Representations}.

\bibitem[{Xiao et~al.(2013)Xiao, Li, Xie, and Tillmann}]{6693084}
Xusheng Xiao, Sihan Li, Tao Xie, and Nikolai Tillmann. 2013.
\newblock \href {https://doi.org/10.1109/ASE.2013.6693084} {Characteristic studies of loop problems for structural test generation via symbolic execution}.
\newblock In \emph{2013 28th IEEE/ACM International Conference on Automated Software Engineering (ASE)}, pages 246--256.

\bibitem[{Yuan et~al.(2024)Yuan, Liu, Ding, Wang, Chen, Peng, and Lou}]{ut_gpt}
Zhiqiang Yuan, Mingwei Liu, Shiji Ding, Kaixin Wang, Yixuan Chen, Xin Peng, and Yiling Lou. 2024.
\newblock \href {https://doi.org/10.1145/3660783} {Evaluating and improving chatgpt for unit test generation}.
\newblock \emph{Proc. ACM Softw. Eng.}, 1(FSE).

\bibitem[{Zhang et~al.(2023{\natexlab{a}})Zhang, Wang, Xia, Wang, and Li}]{zhang2023algo}
Kexun Zhang, Danqing Wang, Jingtao Xia, William~Yang Wang, and Lei Li. 2023{\natexlab{a}}.
\newblock \href {https://openreview.net/forum?id=JolrEmMim6} {{ALGO}: Synthesizing algorithmic programs with generated oracle verifiers}.
\newblock In \emph{Thirty-seventh Conference on Neural Information Processing Systems}.

\bibitem[{Zhang et~al.(2023{\natexlab{b}})Zhang, Yu, Hashimoto, Lewis, Yih, Fried, and Wang}]{zhang2023codereviewer}
Tianyi Zhang, Tao Yu, Tatsunori Hashimoto, Mike Lewis, Wen-Tau Yih, Daniel Fried, and Sida Wang. 2023{\natexlab{b}}.
\newblock \href {https://proceedings.mlr.press/v202/zhang23av.html} {Coder reviewer reranking for code generation}.
\newblock In \emph{ICML}, pages 41832--41846.

\bibitem[{Zhang et~al.(2024)Zhang, Yao, Zhang, Yun, Yu, Li, and Tang}]{zhang-etal-2024-transferable}
Xiaokang Zhang, Zijun Yao, Jing Zhang, Kaifeng Yun, Jifan Yu, Juanzi Li, and Jie Tang. 2024.
\newblock \href {https://doi.org/10.18653/v1/2024.acl-long.668} {Transferable and efficient non-factual content detection via probe training with offline consistency checking}.
\newblock In \emph{Proceedings of the 62nd Annual Meeting of the Association for Computational Linguistics (Volume 1: Long Papers)}, pages 12348--12364, Bangkok, Thailand. Association for Computational Linguistics.

\bibitem[{Zheng et~al.(2024)Zheng, Zhang, Shen, Liu, Lin, Fu, Chen, and Yue}]{zheng2024opencodeinterpreter}
Tianyu Zheng, Ge~Zhang, Tianhao Shen, Xueling Liu, Bill~Yuchen Lin, Jie Fu, Wenhu Chen, and Xiang Yue. 2024.
\newblock \href {https://doi.org/10.18653/v1/2024.findings-acl.762} {{O}pen{C}ode{I}nterpreter: Integrating code generation with execution and refinement}.
\newblock In \emph{Findings of the Association for Computational Linguistics: ACL 2024}, pages 12834--12859, Bangkok, Thailand. Association for Computational Linguistics.

\end{thebibliography}

\clearpage

\appendix
\label{sec:appendix}
\section{License}
We utilize the CodeFeedback-Filtered-Instruction dataset and the training set from TACO as the data sources for generating high-quality unit tests.
Both datasets are distributed under the Apache 2.0 license, which permits users to freely use, copy, modify, and distribute the software for both personal and commercial purposes.

The parameters of \model, along with the corresponding training data, will be made publicly available upon acceptance.
The training data exclusively consists of synthetic code solutions and unit tests, without any personally identifying information or offensive content.
We will release the LLM-generated data and the models fine-tuned on this data under the Apache 2.0 license.

\label{app:license}

\section{More Results of Unit Test Scaling}
\begin{figure*}[t]
  \centering
  \includegraphics[width=\textwidth]{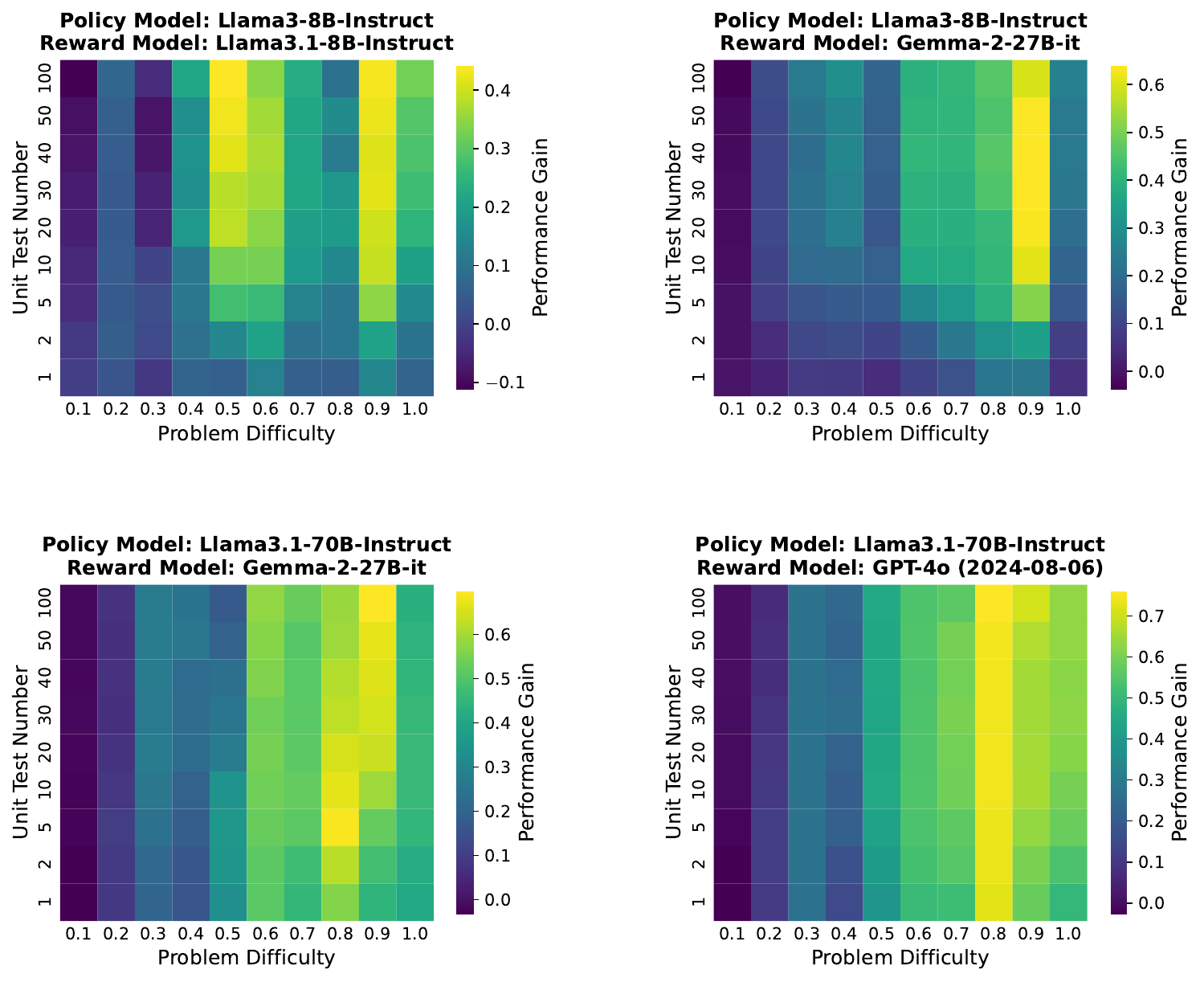}
  \caption{The performance gain of scaling the number of unit tests on problems of different difficulties across various policy model and reward model.
  Overall, increasing the number of unit tests yields greater performance improvements on more challenging problems, particularly when employing Llama3.1-70B as the policy model.}
 \label{fig:app_scale_on_diff_diff_finegrain}
\end{figure*}

Figure~\ref{fig:app_scale_on_diff_diff_finegrain} presents a more fine-grained analysis by categorizing questions based on their difficulty.
The results further confirm that increasing the number of unit tests generally leads to larger performance improvements on more challenging problems.
This trend becomes particularly evident when leveraging more advanced models, such as Llama3.1-70B as the policy model or GPT-4o as the reward model.

\label{app:more_results}

\section{Experiment Settings and Baselines}
\label{app:baseline}
The experiments employ four policy models of various parameter sizes and types, including Llama3-8B-Instruct, Llama3-70B-Instruct, GPT-3.5-turbo, and GPT-4o-mini.
For each policy model, we generate $100$ candidate code solutions, following the hyperparameters in Table~\ref{tab:parameters}.
All models are deployed and inferenced using 8 NVIDIA A800 GPUs.

For the baselines, we first utilize a grading reward model, which takes a candidate code solution as input and outputs a scalar score representing the quality of the solution.
Specifically, we employ ArmoRM-Llama3-8B-v0.1, a powerful reward model that leverages mixture-of-experts (MoE) aggregation across multiple reward objectives.
In addition, we include several baselines that use unit tests as verifiers to identify correct code solutions:

\begin{itemize}
    \item MBR-Exec: This method ranks solutions using minimum Bayes risk (MBR) decoding based on the execution results of LLM-generated test cases. For our experiments, we adopt the hard loss variant of this approach.
    \item CodeT: This baseline evaluates the consistency of outputs with generated test cases and assesses concordance among candidate code solutions through a dual execution agreement mechanism.
    \item MPSC: This method evaluates candidate code solutions from three perspectives: solution, specification, and test case. It constructs a 3-partite graph to identify the optimal solution.
\end{itemize}

To ensure a fair comparison, all baselines leveraging unit tests are provided with the same computational budget of $100$ inferences.
For MBR-Exec and CodeT, we perform $100$ inferences and prompt the LLM to generate $10$ test cases for each inference, following the setup of CodeT.
For MPSC, under the same computational constraints, we instruct the LLM to generate $50$ test cases and $50$ specifications.

\begin{table}[t]
  \setlength{\tabcolsep}{4pt}
  \begin{center}
    \scalebox{0.93}{
    \begin{tabular} {ll}
      \toprule
      Hyperparameters & Value \\
      \midrule
      Temperature & 0.8 \\
      Top P & 0.95 \\
      Frequency Penalty & 0 \\
      Presence Penalty & 0 \\
      \bottomrule
    \end{tabular}
    }
    \caption{The hyperparameters of LLMs for solution and unit test generation.}
    \label{tab:parameters}
  \end{center}
\end{table}

\section{Metrics for Assessing Unit Test Quality}
\label{app:metrics}
We introduce the details for computing the four metrics for evaluating the quality of unit tests in Section~\ref{sec:ut_quality}.
To evaluate the quality of the generated unit tests, we define four metrics: \textit{Accuracy}, \textit{F1 Score}, \textit{False Acceptance Rate (FAR)}, and \textit{False Rejection Rate (FRR)}.
These metrics are computed in the same manner for both individual unit tests and multiple unit tests under the majority voting framework.
However, the interpretation of \textit{True Positives (TP)}, \textit{True Negatives (TN)}, \textit{False Positives (FP)}, and \textit{False Negatives (FN)} slightly differs between these two settings.

The four metrics are formally defined as follows.
\textbf{Accuracy} measures the proportion of correct predictions and is given by:

\begin{equation}
    \text{Accuracy} = \frac{TP + TN}{TP + TN + FP + FN}
\end{equation}

\textbf{Precision} quantifies the proportion of predicted positives that are actually correct:

\begin{equation}
    \text{Precision} = \frac{TP}{TP + FP}
\end{equation}

\textbf{Recall} (also known as True Positive Rate) measures the proportion of actual positives that are correctly identified:

\begin{equation}
    \text{Recall} = \frac{TP}{TP + FN}
\end{equation}

\textbf{F1 Score} is the harmonic mean of precision and recall:

\begin{equation}
    \text{F1 Score} = \frac{2 \cdot \text{Precision} \cdot \text{Recall}}{\text{Precision} + \text{Recall}}
\end{equation}

\textbf{False Acceptance Rate (FAR)} measures the probability of a wrong solution being incorrectly accepted:

\begin{equation}
    \text{FAR} = \frac{FP}{FP + TP}
\end{equation}

\textbf{False Rejection Rate (FRR)} measures the probability of a correct solution being incorrectly rejected:

\begin{equation}
    \text{FRR} = \frac{FN}{FN + TN}
\end{equation}

The definitions of \textit{True Positives (TP)}, \textit{True Negatives (TN)}, \textit{False Positives (FP)}, and \textit{False Negatives (FN)} are as follows.
\textit{True Positives (TP)} denote the number of correct solutions that are classified as correct, while \textit{True Negatives (TN)} refer to the number of incorrect solutions that are classified as incorrect. \textit{False Positives (FP)} represent the number of incorrect solutions that are classified as correct, and \textit{False Negatives (FN)} are the number of correct solutions that are classified as incorrect.

The distinction between a single unit test and majority voting lies in the classification process.
For a single unit test, the classification is directly determined by whether the unit test accepts or rejects a solution.
For multiple unit tests under majority voting, a solution is classified as \textit{correct} if it is accepted by the largest number of unit tests among all candidate solutions, while the remaining solutions are classified as \textit{incorrect}.
These definitions ensure the metrics are consistently applicable across both single and multiple unit test scenarios.

\onecolumn
\section{Prompt for Data Synthetic Pipeline}
\label{app:prompt}

\begin{tcolorbox}[title = {Prompt for Filtering Unsuitable Questions for Unit Testing}]
Below is a programming question and a Python code solution. You need to determine whether this question is challenging to evaluate using traditional unit tests. Apply the following criteria to identify questions that are hard to evaluate using unit tests: \\

1. Functions involving randomness or probability:\\
   1) Random number generators;
   2) Shuffling algorithms;
   3) Probability-based functions\\

2. Time-dependent functions:\\
   1) Functions that get the current time;
   2) Timer functions\\

3. Functions relying on external resources:\\
   1) Network request functions;
   2) File system operations;
   3) Database queries\\

4. Concurrency and multithreading functions:\\
   1) Thread synchronization functions;
   2) Concurrent operation functions\\

5. Hardware-related functions:\\
   1) Device driver functions;
   2) Hardware sensor reading functions\\

6. User interface related functions:\\
   1) Graphics rendering functions;
   2) User input processing functions\\

7. Functions with side effects:\\
   1) Functions modifying global state;
   2) Logging functions\\

8. Cryptography-related functions:\\
   1) Functions generating encryption keys;
   2) Certain encryption algorithm implementations\\

9. Machine learning and adaptive algorithm functions:\\
   1) Model training functions;
   2) Neural network backpropagation algorithms;
   3) Self-tuning algorithms\\

10. Complex mathematical or simulation functions:\\
    1) High-precision floating-point calculations;
    2) Physical simulations (e.g., fluid dynamics, particle collisions);
    3) Complex optimization algorithms\\

\#\#\# programming question\\
\{\textcolor{purple}{question}\}\\

\#\#\# code solution\\
```python\\
\{\textcolor{purple}{code}\}\\
```\\

Let's think step by step: If the question and answer meet the above criteria, please answer YES; otherwise, answer NO. Please first give the reason for your judgments, followed by your decision. Your decision should be in the last line of the reply, which ONLY contains one word: YES or NO.\\
\end{tcolorbox}

\begin{tcolorbox}[title = {Prompt for Unit Test Repairation}]
I currently have an incorrect unit test code, where some of the output does not match the correct answer. After running the unit test on the correct code answer, I obtained the execution result of the unit test. Please modify the original unit test based on the execution result to make the output correct. \\

\#\#\# Unit test with error output\\
\{\textcolor{purple}{unit\_test}\}\\
\#\#\# Execution result with the ground truth\\
\{\textcolor{purple}{exec\_result}\}\\

You should output the complete modified unit test code in markdown format. Do NOT add any additional comments to the original unit test except for modifying the output.\\
\end{tcolorbox}

\begin{tcolorbox}[title = {Prompt for Code Entrance Generation}]
Below is a code solution and a corresponding unit test. Please locate the function in the code solution that the unit test is testing, and output the function name, return values, and input parameters. \\

\#\#\# Code Solution\\
\{\textcolor{purple}{code}\}\\
\#\#\# Execution result with the ground truth\\
\{\textcolor{purple}{unit\_test}\}\\

Please output the function name, return values, and input parameters in the following format WITHOUT any other words:\\
Function name:\\
Input parameters:\\
Return values:\\
Function declaration:\\
\end{tcolorbox}

\begin{tcolorbox}[title = {Prompt for Generating Solutions for TACO Dataset}]
Below is a programming question and an answer format. You need to use Python to answer this question following the provided function format. \\

\#\#\# Programming Question\\
\{\textcolor{purple}{question}\}\\
\#\#\# Answer Format\\
\{\textcolor{purple}{answer\_format}\}\\

Please follow the answer format to output your answer in markdown format. You need to return the values required by the question instead of printing them. Attention: You ONLY need to output code answer in function format WITHOUT any other words.\\
\end{tcolorbox}

\begin{tcolorbox}[title = {Prompt for Genearting Unit Tests for Reward Model}]
Below is a question and it's corresponding code answer. Please write test cases to check the correctness of the code answer. You need to use the unittest library in Python and create a test class for testing. \\

\#\#\# question\\
\{\textcolor{purple}{question}\}\\
\#\#\# code solution\\
\{\textcolor{purple}{code}\}\\

Please add detailed comments to the test cases you write. You do not need to test the function's ability to throw exceptions.\\
\end{tcolorbox}

\begin{tcolorbox}[title = {Prompt for Generating Code Solutions for Policy Model}]
Please provide a self-contained Python script that solves the following problem in a markdown code block: \\
\lstinline{```}Python\\
\{\textcolor{purple}{prompt}\}\\
\lstinline{```}\\
\end{tcolorbox}

\begin{tcolorbox}[title = {Prompt for Reorganizing Questions in CodeFeedBack-Filtered Dataset}]
Below is a programming question and its corresponding answer in Python code. Please reorganize the programming question to make it precise and clear. The reorganized question should only contain the required function name, information, and restriction of the question and exclude any irrelevant information and irrelevant code snippets. ONLY output the reorganized question in one or few paragraphs without headline, title, subtitle, etc. \\

\#\#\# programming question\\
\{\textcolor{purple}{query}\}\\
\#\#\# answer in Python code\\
\{\textcolor{purple}{answer}\}\\
\end{tcolorbox}

\begin{tcolorbox}[title = {Prompt for Reorganizing Code Solutions in CodeFeedBack-Filtered Dataset}]
Below is a programming question and its corresponding answer in Python code. Please reorganize the answer to include ONLY the required function in the code snippets. The reorganized answer should exclude any test case in the original code snippets and ensure the function name (entance) is consistent with the name in the question. \\

\#\#\# programming question\\
\{\textcolor{purple}{query}\}\\
\#\#\# answer in Python code\\
\{\textcolor{purple}{answer}\}\\
\end{tcolorbox}

\twocolumn

\begin{figure*}[t]
  \centering
  \includegraphics[width=0.95\textwidth]{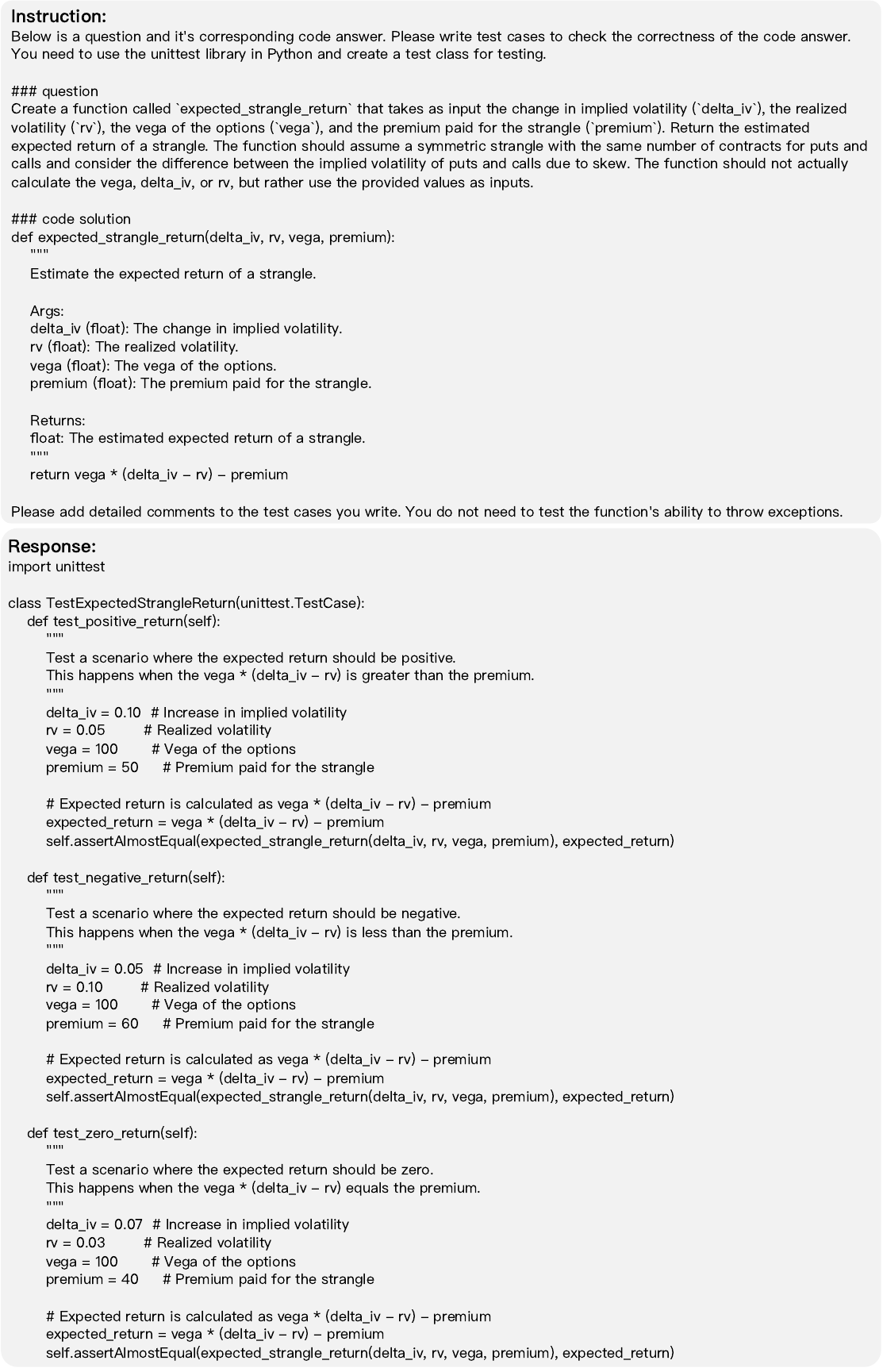}
  \caption{An example of the training data for the unit test generator.}
 \label{fig:data_example}
\end{figure*}

\end{document}